\documentclass[10pt,twocolumn,letterpaper]{article}

\usepackage{iccv}
\usepackage{times}
\usepackage{epsfig}
\usepackage{graphicx}
\usepackage{amsmath}
\usepackage{amssymb}
\usepackage{booktabs}
\usepackage{multirow}
\usepackage[dvipsnames]{xcolor}
\usepackage{soul}
\usepackage[normalem]{ulem}

\usepackage[breaklinks=true,bookmarks=false]{hyperref}

\iccvfinalcopy 

\def\iccvPaperID{****} 

\ificcvfinal\fi

\begin{document}

	\title{Semantic Compositional Learning for Low-shot Scene Graph Generation}
	
\author{Tao He$^1$, Lianli Gao$^2$, Jingkuan Song$^2$, Jianfei Cai$^1$,  Yuan-Fang Li$^1$ \thanks{Corresponding author.} \\ 
	$^1$Faculty of Information Technology, Monash University \\
	$^2$Center for Future Media, University of Electronic Science and Technology of China\\
	{\tt\small \{tao.he,yufang.li\}@monash.edu, lianli.gao@uestc.edu.cn,jingkuan.song@gmail.com} 
}

	\maketitle
	\ificcvfinal\thispagestyle{empty}\fi
	
	\begin{abstract}
		Scene graphs provide valuable information to many downstream tasks. Many scene graph generation (SGG) models solely use the limited annotated relation triples for training, leading to their underperformance on low-shot (few and zero) scenarios, especially on the rare predicates. To address this problem, we propose a novel semantic compositional learning strategy that makes it possible to construct additional, realistic relation triples with objects from different images. 
		Specifically, our strategy decomposes a relation triple by identifying and removing the unessential component, and composes a new relation triple by fusing with a semantically or visually similar object from a visual components dictionary, whilst ensuring the realisticity of the newly composed triple. 
		Notably, our strategy is generic, and can be combined with existing SGG models to significantly improve their performance. 
		We performed comprehensive evaluation on the benchmark dataset Visual Genome.  For three recent SGG models, adding our strategy improves their performance by close to 50\%, and all of them substantially exceed the current state-of-the-art.

	\end{abstract}

	\section{Introduction} \label{intro}
	
	Scene Graph Generation (SGG) is a fundamental task for many downstream applications in computer vision, including visual captioning~\cite{yao2018exploring,yang2019auto} and visual question answering~\cite{teney2017graph,antol2015vqa}, as SGG can provide those tasks with a specific relation graph of an image, which makes it possible for the downstream models to conduct high-level graph reasoning over the generated scene graph and further capture more comprehensive information. 
	Generally, a scene graph (SG) can be represented as a set of relation triples, i.e.\ $<$subject, predicate, object$>$, or simply $<$s, p, o$>$. 
	
	\begin{figure}
		\centering
		\includegraphics[width=0.99\linewidth]{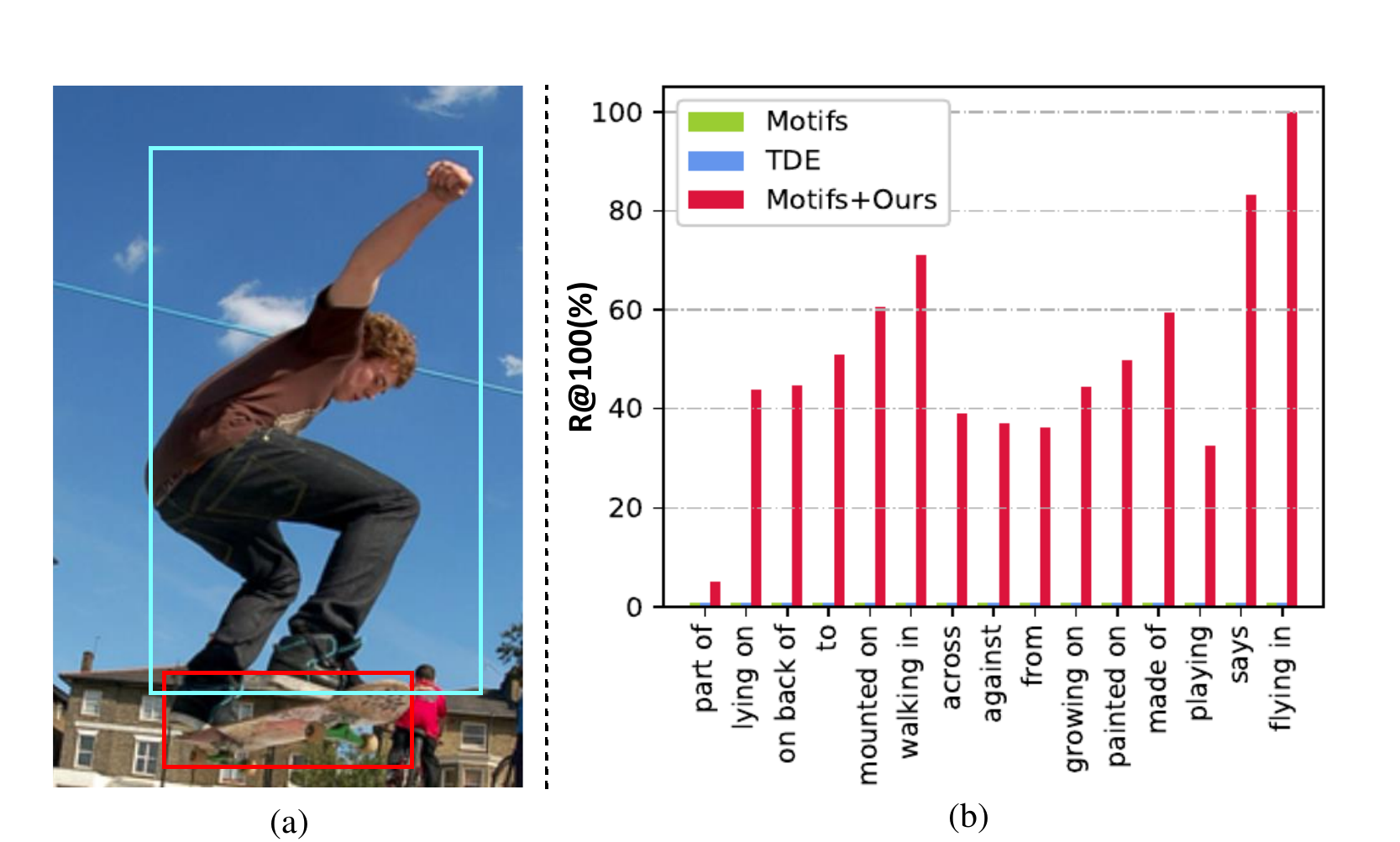}
		
		\caption{ (a) An example of common relation, i.e.\ $<$man, standing on, skateboard$>$, where the man and skateboard have different impacts on their relationship \texttt{standing on}. (b) Performance (as measured by R@100) of our proposed method compared with other two models: Motifs~\cite{zellers2018neural} and TDE~\cite{tang2020unbiased} on the $15$ rarest predicates on the task of PredCls.}
		\label{fig:1}
	\end{figure}
	
	{It has been observed that in real-world datasets, some predicates, e.g. \texttt{flying in}, are extremely rare. 
		Moreover, the number of possible relations grows as the product of the number of subjects, objects and relations. With hundreds of subjects/objects in large-scale datasets, there can be millions of possible relations. 
		These factors make the scene graph  generation more challenging.} 
	Specifically, as shown in a recent study~\cite{zellers2018neural}, in the widely-used Visual Genome dataset~\cite{krishna2017visual}, there are $150$ object categories and $50$ distinct relations. Thus, the total number of combinations of the relation triples is more than one million, while the total number of training images is only about $50,000$. 
	This fact leads to the arduous challenge that the majority of relation triples may only have a few samples, although the corresponding predicate has many samples. For example, in Visual Genome, the predicate \texttt{sitting on} {belongs to}  a  frequent predicate with more than $3,000$ samples. However, if we focus on the number of samples on each relation triple, it can be observed that the majority of the samples belong to $<$man, sitting on, chair$>$, but only several exemplars are for $<$man, sitting on, car$>$. Therefore, even though a model may classify the former well, it may fail on the 
	rare relation triples,  e.g. $<$man, sitting on, car$>$. 
	
	Furthermore, we also observe that given a relation triple, in many cases the impact of the subject and the object on the predicate is different. For example, as shown in Figure \ref{fig:1}(a), even though we remove the skateboard from the image, we could still guess the predicate is likely to be \texttt{standing on}, because the \texttt{standing on} is more related to the subject instead of the object. I.e., the pose of the subject dominantly determines the predicate, while the visual appearance feature of the object makes little difference. 
	Based on this observation, we devise a framework that  first \emph{decomposes} the  representation of a relation triple into two parts: the important and  the unimportant, and simultaneously   assign  their correlation on the predicate, and then \emph{compose} a new relation by fusing the important element with another potential object. 
	Specifically, the benefits of doing so are threefold. 
	
	First, for those rare relations, our method can generate sufficient samples for effective learning. Taking the rare relation triple $<$man, sitting on, car$>$ as an example, it is obvious that the object \texttt{car} plays an unessential role, and thus when we change the car instance with other car instances,   the predicate is still \texttt{sitting on}. 
	Fortunately, the Visual Genome dataset has many car instances with a wide range of visual appearance features, including colors, shapes and textures. 
	Therefore, by the decomposition and composition operations, we can generate many additional triples for the relation $<$man, sitting on, car$>$, which alleviates the data-starvation problem for those rare relation. 
	
	Second, our strategy can address the challenging task of  zero-shot relation recognition. Specifically, our method can compose novel relation triples by fusing  the essential element with an instance from a semantically similar object category. For instance, in Visual Genome,   there is no relation triple $<$man, sitting on, train$>$, but  in practice,  although this relation rarely occurs, it indeed exists in some movies. Most of existing SGG models~\cite{zellers2018neural,tang2020unbiased,tang2019learning,yang2018graph,xu2017scene} are likely to predict this predicate as a non-relation, as those models have not seen this relation triple in the training set. In our method,  we first decompose the triple $<$man, sitting on, car$>$ into two parts: man and car, and then compose novel relation triples by fusing the essential man with an instance in semantically similar object categories, e.g.\ train and boat, both of which belong to the vehicle category. After that, our model can be trained with those composed, unseen triples and thus address the zero-shot relation detection problem. 
	
	Last, our strategy can be seamlessly integrated with existing SGG models, as our strategy only concerns with relation construction, and other feature learning strategies such as~\cite{zellers2018neural,zhang2017visual,Lin_2020_CVPR} can be reused. Figure~\ref{fig:1}(b) highlights the effectiveness of our model on rare predicates in terms of R@100. It plots the performance comparison result of our proposed method and two competitive models on the 15 rarest relations in Visual Genome, where TDE~\cite{tang2020unbiased} is the previous best model. It can be obviously seen that both of the compared methods handle the rare relations poorly (close to 0 R@100 values), while our model performs significantly better on all of these relations. 
	
	In summary, our main contributions are threefold:
	\begin{itemize} 
		\item We first propose a novel decomposition and composition strategy for Scene Graph Generation, which addresses the critical low-shot and zero-shot relation detection problems. Furthermore, our method can be seamlessly plugged in other SGG models and significantly improve their performance. 
		
		\item We develop a simple and effective method to calculate the correlations of the subject and object to their predicate to retrieve the best matched visual components.
		
		\item We evaluate our model on the standard tasks of relation retrieval,  few-shot and zero-shot settings. In all three settings, our model significantly surpasses state-of-art methods. Notably, our model achieves an improvement of nearly 50\% compared with them. 
	\end{itemize}

	\section{Related Work}
	\noindent\textbf{Scene Graph Generation (SGG)} is a task to detect visual relationships, which requires the model to localize and recognize them. The main challenge lies in the variability of   combinations of  subjects and objects, leading to the data-starving issue for the majority of relationships.  In the past, most of SGG 
	methods predominantly focused on refinement of object and relation features early on~\cite{xu2017scene,lu2016visual,newell2017pixels,he2020learning,dai2017detecting,herzig2018mapping}, while consistently ignoring an important property of the Visual Genome (VG) dataset, that its relation distribution is long-tail. This biased distribution was first revealed in Motifs~\cite{zellers2018neural}, which statistically showed the frequency of each relation. Interestingly, it was observed that when only using the statistical frequency information as prior knowledge to classify the relations, that is, without considering any visual feature, Motifs achieves competitive results, of only about 1.5\% lower than state-of-the-art methods~\cite{xu2017scene,lu2016visual}. Inspired by Motifs, many recent works~\cite{zareian2020bridging,tang2019learning,tang2020unbiased,chen2019knowledge} invested heavily into solving the data bias or long tail problems in SG datasets.  However, those methods still fail when facing those highly rare or unseen relation triples.  
	
	\noindent \textbf{Compositional Learning} has received tremendous attention in many visual tasks~\cite{burgess2019monet,kato2018compositional,zhu2016unconstrained}. The core idea of compositional learning is to use the limited samples to compose additional exemplars that maintain the main semantic meanings. 
	Christopher et.al.~\cite{burgess2019monet} proposed a compositional generative model, named as Multi-Object Network (MONet), to decompose scenes into abstract building blocks by several Variational Auto-Encoder networks. For the multi-label classification problem, Alfassy et.al.~\cite{alfassy2019laso} proposed a feature compositional network to generate new multi-labels by simulating set operations.
	Kato, Li and Gupta~\cite{kato2018compositional} proposed a novel framework which incorporates an external knowledge graph and then uses graph
	convolutional networks to compose classifiers for objects pairs. Hou et. al.~\cite{hou2020visual} proposed a compositional learning strategy for the task of Human-Object Interaction (HOI) detection by composing new interaction  samples in the feature space.  {Peyre et.al.~{\cite{peyre2019detecting}} used analogies to learn the compositional representations for subjects, predicates, and objects and during testing stage, they use the nearest neighbour search to retrieve the similar triples. } 
	
	All of the  three \cite{kato2018compositional,hou2020visual,peyre2019detecting} works attempt to solve the zero-shot learning problem in HOI detection by composing unseen interactions. Our idea is different from them in two important aspects. Firstly, the subject in  interaction triples of Human-Object Interaction is fixed, which leads to the fact that existing methods do not consider the impacts of the subject on the verbs (i.e.\ predicates) and only need to change the object. Therefore, the composition for scene graph relation is much more challenging that interactions in HOI. Secondly, existing works only consider the inter-class composition and consistently ignore the intra-class composition, leaving great potential improvement room. 
	

	\section{Methodology} \label{meth}
	\begin{figure*}
		\centering
		\includegraphics[width=0.92\linewidth]{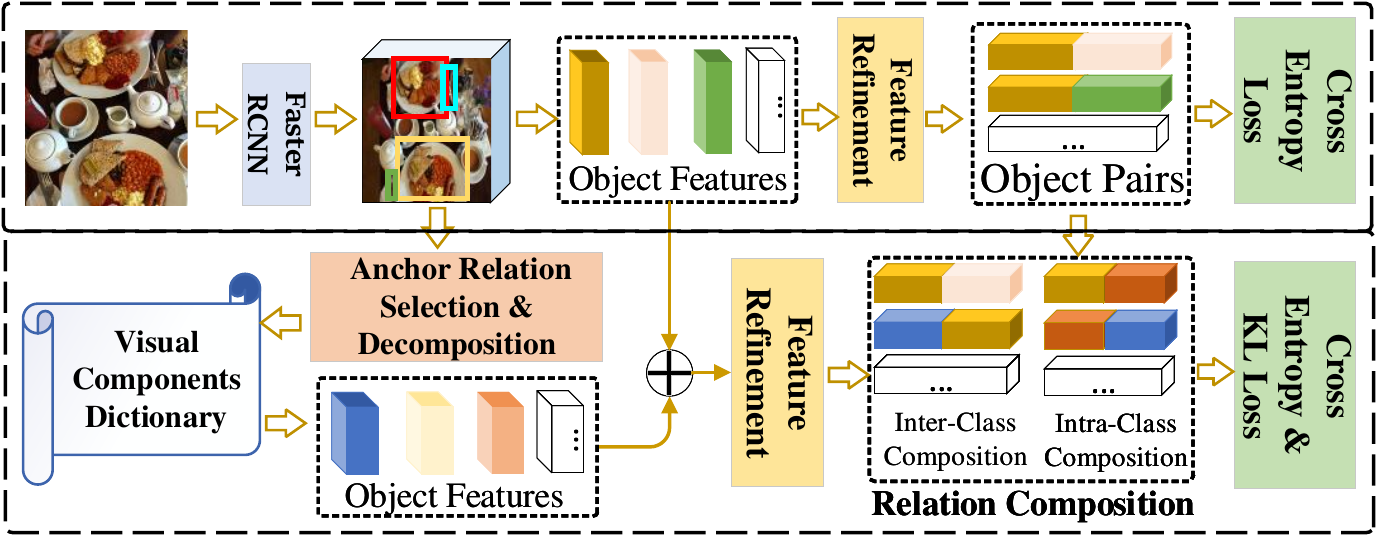}
		\caption{The overall architecture of our framework, where the top block denotes general Scene Graph Generation models, such as Motifs~\cite{zellers2018neural}, TransE~\cite{zhang2017visual} and GPSN~\cite{Lin_2020_CVPR}, while the bottom pipeline denotes our proposed compositional learning method for SGG.  
		}
		\label{fig:frameword}
	\end{figure*}
	
	Figure~\ref{fig:frameword} illustrates our overall framework, where  the top half generally depicts convention SGG models and the bottom half shows our proposed compositional strategy. 
	Our strategy can be divided into two stages: (1) anchor relation 
	selection and decomposition, and (2) relation composition. 
	Specifically, the first stage aims at selecting anchor relation triples, i.e. those that can be decomposed, and then decomposing them into two parts: essential and unessential. Then, we utilize a similarity measurement to find the best matched candidate from a visual components dictionary generated by an object detection network. In the second stage, we use the anchor relation and those candidates  to compose new relations with inter- and intra-class  instances. 
	
	
	\subsection{Revising Scene Graph Generation Models}
	Generally, we could characterize most of SGG models~\cite{zellers2018neural,zhang2017visual,xu2017scene,tang2019learning,Lin_2020_CVPR} in three main steps, as shown the top part of Figure~\ref{fig:frameword}. The first step is to use an off-the-shelf pre-trained object detection network, e.g.\ Faster-RCNN~\cite{ren2015faster}, to extract spatial, semantic and visual object features. Next, they deploy a context learning module to refine those features, e.g.  surrounding context~\cite{zellers2018neural} or edge direction context~\cite{Lin_2020_CVPR}. 
	Finally, they construct subject-object pairs from the same image and deploy a predicate classifier to classify those object pairs. 
	Formally, we could denote the three steps with the below formula:
	
	\begin{equation}
	\mathbf{p} = cls(\mathcal{C}(\mathcal{F}(\mathcal{D}(\mathrm{I}))))
	\label{eq.1}
	\end{equation}
	where $\mathcal{D(.)}$ is the object detection network, aiming at detecting visual components in the image $\mathrm{I}$, i.e object proposals. $\mathcal{F}(.)$ is the feature refinement module, such as \cite{zellers2018neural,xu2017scene,Lin_2020_CVPR}. $\mathcal{C}(.)$ is a {subject-object pairs construction function}    aiming at fusing two object features into a relation representation. $cls(.)$ is a final relation classifier and $\mathbf{p}$ is the predicted relation  score. 
	
	As mentioned before, the majority of existing works put their efforts in  the modification of the feature refinement module $\mathcal{F}$. However, a common and important issue of those works is that they only use the detected objects in the same image to compose the relation triples but ignore the fact that even two objects from different images can also compose a relation triple. Based on this  observation, chiefly, we could express our idea as the formula below:
	
	\begin{equation}
	\mathbf{p} = cls(\mathcal{C}^{'}(\mathcal{F}( \mathcal{S}( \mathcal{V}))))
	\label{eq.2}
	\end{equation}
	where { $\mathcal{V} $ denotes {  visual components dictionary of training set} detected by $\mathcal{D}$}; $\mathcal{S}$ is a selection function to choose suitable anchor relation triples and visual components from all candidates; and $\mathcal{C}^{'}(.)$ is a composition function but different from $\mathcal{C}(.)$ in Eq.~\ref{eq.1}. In the following subsections, we will describe each of them.
	
	\subsection{Visual Components Detection}
	
	The first step is to generate all the visual components of the dataset. Following most SGG models~\cite{zellers2018neural,Chen_2019_CVPR},  we also use the pre-trained Faster-RCNN~\cite{ren2015faster} as the backbone object detection network. Subsequently,  a region proposal network (RPN) is deployed to generate a set of $n$ bounding boxes $\mathcal{B}_i=\{\mathbf{b}_1,\mathbf{b}_2,\ldots, \mathbf{b}_n\}$ for each image $i$, where $\mathbf{b}_j = [x^j_t, y^j_t, x^j_b, y_b^j ]$, $(x^j_t,y^j_t)$ denotes the top-left coordinate and $(x^j_b,y^j_b)$ is the bottom-right coordinate. The corresponding object visual features are represented as $\mathcal{O}_i=\{\mathbf{f}^o_1,\mathbf{f}^o_2,\ldots,\mathbf{f}^o_n\}$, aligned by a RoIAlign module~\cite{he2017mask}, where $\mathbf{f}^o_j \in \mathbb{R}{^{4096}} $ for each $j$. Additionally, we extract the localisation feature of the object proposals  and transform each bounding box $\mathbf{b}_j \in \mathbb{R}{^{4}}$ into a $128$-dimensional informative spatial feature vector, denoted as $\mathbf{f}^s_j  \in \mathbb{R}{^{128}}$. 
	Previous work~\cite{zellers2018neural,tang2020unbiased} have shown that the language prior knowledge can serve as auxiliary information to enrich the object features. Therefore, we further extract the word embedding features of the object label $l_j$ as $\mathbf{f}^w_j \in \mathbb{R}^{200}$. Finally, we concatenate  the  three types of features into a comprehensive vector as the visual component, i.e. $\mathbf{f}^i_j=\left[\mathbf{f}^o_j;\mathbf{f}^s_j;\mathbf{f}^w_j\right] \in \mathbb{R}^{4424}$. Thus, the visual component dictionary $\mathcal{V}$ is  denoted as $\{ \{ \mathbf{f}^1_j\}^n_{j=1},\{\mathbf{f}^2_j\}^n_{j=1} \cdots \{\mathbf{f}^m_j\}^n_{j=1}\}$, where $m$ is the number of images.

	\subsection{Anchor Relation  Selection and Decomposition}
	
	Our overall goal is to decompose a real relation and compose a new relation by fusing with another visual component.
	Theoretically, we could formulate the goal as the below:
	\begin{equation}
	\begin{array}{ll}
	& f(r') \approx f(r), where:   \\
	& f(r') = f(r)\ominus f(u) \oplus f(u')
	\end{array}
	\label{eq.rc}
	\end{equation}
	where $f$ is a feature learning function; and $\ominus$ denotes the decomposition operation whereas $\oplus$ is the composition operation. $f(r)\ominus f(u) \oplus f(u')$ is the composed relation triple, $u$ is the unessential object in the anchor relation $r$ and $u'$ is the other visual component from $\mathcal{V}$. The selection of $u'$ is described in the next section. 
	
	\noindent \textbf{Anchor Relation Selection.}  
	As analysed before, we observe that in some relation triples, the subject and object have different impacts on their predicate, and removing the unessential one does not make a difference for their relationship. We name those relation triples that satisfy this property as \emph{Anchor Relations}. 
	Not all relation triples are anchor relations.  Intuitively, if two objects in a relation triple {elaborate}, they  should have less shared information, because much shared information means that both objects have a large impact on their relationships and their interaction  is relatively tight, which makes it challenging to decompose or disentangle them. That is, $f(r)\ominus f(u)$ could be too far away from $f(r)$ and consequently Eq.\ \ref{eq.rc} becomes hard to satisfy. On the other hand, the shared information can be an important clue   for some predicates,   but  the decomposition operation could break this clue.   For the selection of anchor relations, we need to  choose those that the  subject and the object are less {entangled}  Therefore, if   boxes of a subject and an  object in a relation have a smaller intersected region, i.e., a smaller $\mathrm{IoU}$ score, they have less common information. 
	
	\noindent\textbf{Anchor Relation Decomposition.}
	This operation aims at determine, given an anchor relation $r$, which of the subject or the object, denoted as $u$ in Eq. \ref{eq.rc}, should be decomposed from the anchor relation. 
	Naturally,  the smaller object is usually inconsequential in most cases, because its features have less contribution to their union feature. Furthermore, the variety of small objects, such as \texttt{bowl} and \texttt{fork},  are  much smaller  than the big objects, e.g.\ \texttt{man}. Therefore, we choose the smaller object as the decomposed element. 
	
	In summary,  given a triple $<\!\!s, r, o\!\!>$, we formulate the selection and decomposition   rules  as follows:
	\begin{equation}
	\mathcal{S}(s,o)\! =\! \left\{ \!\!
	\begin{array}{l}
	\!s\!  ~~~ \rm{if}~  {\mathrm{IoU}}(\mathbf{b}_s,\mathbf{b}_o)\!\! < \!\!\delta ~{\&}~\mathrm{A}(\mathbf{b}_s) \!\!<\!\! \mathrm{A}(\mathbf{b}_o)  \\
	\!o\! ~~~  \rm{if}~ \mathrm{IoU}(\mathbf{b}_s,\mathbf{b}_o)\!\! <\!\! \delta ~{\&}~\mathrm{A}(\mathbf{b}_s)\!\! >\!\! \mathrm{A}(\mathbf{b}_o)     \\
	\!-1\!  ~~~ \mathrm{otherwise}
	\end{array} \right.
	\label{eq:replace}
	\end{equation}
	where   $s$ and $o$ denote the subject and object of a ground-truth relation triple; $-1$ means the relation is not an anchor relation, otherwise $\mathcal{S}(.)$ returns the unessential element.
	$\mathrm{A}(\mathbf{b}_s)$ is an area calculation function for a bounding box $\mathbf{b}_s$; $\mathrm{IoU}(\mathbf{b}_s,\mathbf{b}_o)$ returns the $\mathrm{IoU}$ score of two bounding boxes $\mathbf{b}_s$ and $\mathbf{b}_o$; and $\delta$  is a threshhold for the $\mathrm{IoU}$ score. In our experiments, we empirically set it to $0.3$.  
	
	\subsection{Relation  Composition}
	In this stage, we aim at composing new relation triples based on  anchor relations, where the key challenge lies in how to choose the best appropriate visual component, {i.e. $u'$ in Eq.~\ref{eq.rc},} to fuse  with the anchor relation so that the composed relation is semantically and visually  close to the {anchor relation}.  
	As illustrated in Eq.~\ref{eq.rc}, our aim is to force the representation of the  composed relation,  i.e.\ $f(r')$ in Eq.~\ref{eq.rc}, to be close to the anchor relation $f(r)$. 
	
	More concretely, we devise two types of composition strategies, intra- and inter-class   compositions,  according to whether or not the object labels of $u$ and $u'$ are same.   For the intra-class composition, we require both of $u$ and $u'$ belong to the same object category, which can solve the problem of low-shot relation classification problem, while the inter-class composition could compose novel relation triples not present in the training dataset, which helps the model tackle the zero-shot relation classification problem.  
	
\begin{figure}
		\centering
		\includegraphics[width=1\linewidth]{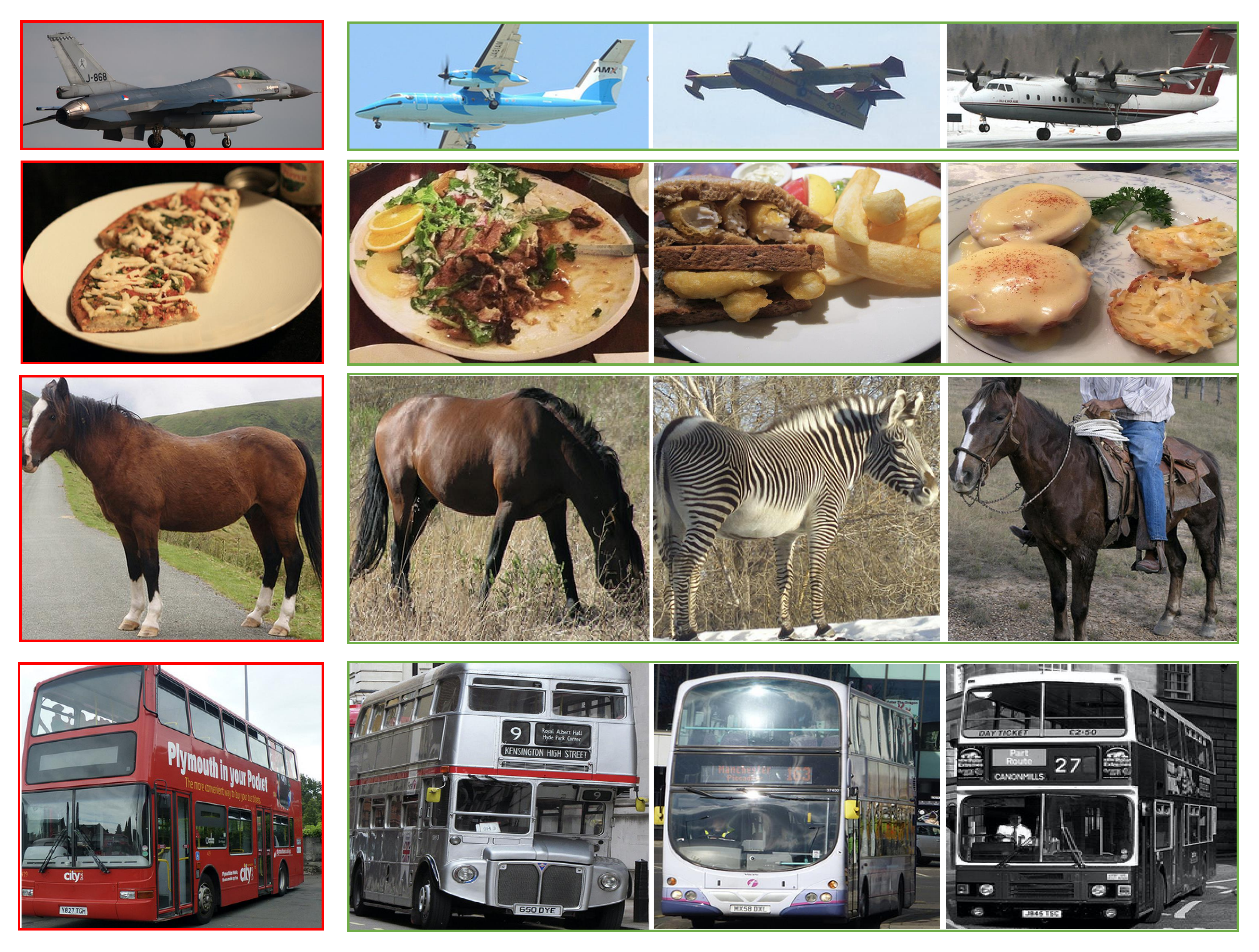}
		\caption{ A few examples of the retrieved similar visual components, where the first column denotes the query component, i.e. $u$ in Eq. \ref{eq.rc} and the right components are the corresponding similar objects based on the similarity measurement of Eq. \ref{eq.ra}.
		}
		\label{fig:re}
	\end{figure}

	\noindent\textbf{Intra-Class Composition.}  It is intuitive that   the distance of two objects' visual  features is a good way to measure two same-category objects' similarity, but the critical drawback lies in time efficiency. As object features are $4096$-dimensional vectors, it is computationally infeasible to calculate the distance for every $u'$ across the entire set of visual components. 
	As an alternative, we design a simple but effective metric on the shapes of two bounding boxes, as we deem that the shape of objects can be an important cue to distinguish them from each other. The main reasons are twofold. First, for the majority of static objects, such as dishes and bottles, their shape and pose does not change a lot, and thus they are visually similar if they have similar shapes. Second, for those dynamic object, e.g., persons or animals, to some extend, their shape information can imply their visual appearance features. It could be argued that the shape information may not be quite precise at representing these dynamic objects. However, since these objects always play an important role in the predicate, we usually do not decompose them, i.e. $u$ is less likely to be a dynamic object. Figure~\ref{fig:re} gives some examples $u'$ retrieved by the shape of $u$. 

	Formally, given two bounding boxes $\mathbf{b}_u$, $\mathbf{b}_v$   of $u$ and $u'$ respectively, we first normalize their  top left corners    to $(0,0)$, as below: 
	\begin{align}
	\mathbf{b}^{'}_u &= [0, 0, x^u_b-x^u_t, y^u_b-y^u_t]\\
	\mathbf{b}^{'}_v &= [0, 0, x^v_b-x^v_t, y^v_b-y^v_t]
	\end{align}
	Then, we use the overlapping of two boxes to measure their similarity:
	\begin{align}
	\mathrm{InS}(\mathbf{b}^{'}_u,\mathbf{b}^{'}_v)&=\mathrm{min}(x^{u'}_b,x^{v'}_b) \cdot \mathrm{min}(y^{u'}_b,y^{v'}_b) \\
	\mathcal{M}(\mathbf{b}^{'}_u,\mathbf{b}^{'}_v)&=  \frac{\mathrm{InS}(\mathbf{b}^{'}_u,\mathbf{b}^{'}_v)}{x^{u'}_b\cdot y^{u'}_b+x^{v'}_b\cdot y^{v'}_b - \mathrm{InS}(\mathbf{b}^{'}_u,\mathbf{b}^{'}_v)}\label{eq.ra}
	\end{align}
	where $x^{u'}_b$ and  $x^{v'}_b$ are the normalized coordinates; $\mathrm{InS}$ calculates the overlap of two boxes; and $\mathcal{M}$ calculates two boxes' shape similarity. 
	
	\noindent \textbf{Inter-Class Composition.}  Different from the above, this composition operator does not require $u$ and $v$ have the same object category. Thus, we could compose unseen relation combinations, e.g.\ $<$man, sitting on, train$>$ could derive {from the anchor relation $<$man, sitting on, car$>$}. The key issue in this module is that we need to guarantee the composed relation is reasonable, for example, $<$man, sitting on, eye $>$ is highly unrealistic. To solve this issue, we first use the language prior, i.e.\ the word embedding of object category of $u$ to retrieve the most semantically similar object categories. For example, if $u$ belongs to car, the retrieved object categories are likely to be train, plane or bike, as all of them are vehicles, therefore their language prior features are close. When we obtain the feasible object categories, we use the same similarity measurement in intra-class composition to obtain the best appropriate visual components.
	%
	
	\subsection{Training and Inference}
	\noindent  \textbf{Training.} We use the Cross Entropy (CE) loss to optimize Eq.~\ref{eq.2}, and Kullback-Leibler (KL) loss to optimize Eq.~\ref{eq.rc} to force the distribution of composed relations not to be far from their anchor relations. It is worth noting that $\mathcal{F}$ in Eq.~\ref{eq.2} can be any feature refinement module, such as Motifs~\cite{zellers2018neural}, TransE~\cite{zhang2017visual}, or GPS-Net~\cite{Li_2020_CVPR}, etc. 
	
	In addition,  our method is able to compose many extra relation triples whose labels depend on the anchor relation. Thus, to balance predicates in every training batch, we utilize a balanced sampling strategy to sample  images. Specifically, we firstly randomly select $N$ predicates from all predicates for each training batch 
	and then sample $K$ images for each sampled predicate. Therefore, the batch size is equal to $N\times K$. The benefit  is that each predicate can be equally sampled, avoiding oversampling for frequent predicates and undersampling for the rare. 
	
	\noindent \textbf{Inference.} During testing, we discard the composition strategy, and use the standard SGG models, as defined in Eq.~\ref{eq.1}, to generate scene graphs. 

	\section{Experiments}
	In this section, we evaluate our model on the task of relation retrieval (\textbf{RR}), as well as its two more challenging settings, namely few-shot and zero-shot relation retrieval (\textbf{FR} and \textbf{ZR} respectivly). Furthermore, we discuss the effectiveness of each component in an ablation study. More experimental settings, results, discussions, a qualitative analysis, as well as source code can be found in the supplementary materials.
	
	\subsection{Dataset and Baselines}
	\textbf{Dataset}. We conduct our experiments on the most widely-used and challenging benchmark dataset Visual Genome (VG)~\cite{krishna2017visual}. Following the widely adopted split~\cite{xu2017scene,zellers2018neural,tang2020unbiased,wang2020tackling}, we choose the most frequent $150$ object classes and $50$ predicates to generate scene graphs, with $57,723$ images for training and $26,443$ images for testing. Additionally, $5,000$ images make up the validation set to select the best model and to finetune model parameters. 
	
	\begin{table*}[ht] 
		\centering
		
		\resizebox{1.7\columnwidth}{!}{
			\begin{tabular}{lccccccccc}
				\toprule 
				\multirow{2.5}{*}{Method} & \multicolumn{3}{c}{PredCls} & \multicolumn{3}{c}{SGCls}  &  \multicolumn{3}{c}{SGDet}  \\  \cmidrule(lr){2-4}\cmidrule(lr){5-7}\cmidrule(lr){8-10}
				&   mR@20& mR@50& mR@100& mR@20& mR@50& mR@100 & mR@20 & mR@50 & mR@100  \\\midrule

				KERN~\cite{chen2019knowledge}  &  13.5 & 17.9 & 19.1 & 8.1  & 9.8 & 10.2 &  5.1 & 6.2 & 7.4\\ 

				GLAT~\cite{zareian2020learning}  &  - & 17.8 & 19.3 &- & 9.9 & 10.4 & - & - & - \\ 
				
				GB-Net$^{\dagger}$~\cite{zareian2020bridging}  & 16.3 & 22.1 & 24.0 & 11.4 & 12.7 & 13.4 & 6.6  & 7.1 & 8.5\\ 
				
				TDE$^\dagger$~\cite{tang2020unbiased}  & 17.3 & 25.1 & 28.2 & 9.3 & 13.0 & 14.7 & 6.2 & 8.5 & 10.7\\

				\midrule

				TransE~\cite{zhang2017visual}  &   13.6 & 17.6 & 19.4 & 6.7  &  8.1 & 8.9 &  5.5 & 7.0 & 8.2\\ %
				TransE+\textbf{DeC}& {{23.2}} & {{31.4}} & {{33.1 }} & {12.8} & {13.9} & {16.5} & {8.7} & {10.6}& {12.7}\\ \cmidrule{1-10} 
				
				GPSN$^{\dagger}$~\cite{Lin_2020_CVPR}  &  15.4 & 20.2  &  22.8 &  10.2 & 11.2  &  12.6 & 7.2  & 8.4  & 9.8\\
				GPSN+\textbf{DeC}& {{26.0}} & {{35.9}} & {{38.4 }} & {14.3} & {17.4} & {18.5} & \textbf{10.4} & {11.2}& {15.2}\\ \cmidrule{1-10} 
				
				Motifs~\cite{zellers2018neural}  & 10.2  &13.5 & 14.8 & 6.8 & 7.3 & 7.8 &  4.5 & 5.7 & 6.8\\ 
				Motifs+\textbf{DeC} & \textbf{{26.2}} & \textbf{{35.7}} & \textbf{{38.6 }} & \textbf{15.0} & \textbf{18.4} & \textbf{19.1} & {10.1} & \textbf{13.2}& \textbf{15.6}\\ 
				\bottomrule
			\end{tabular}
		}
		\caption{ 
			Comparison with the state-of-the-art SGG models in terms of mean Recall@K (mR@K). $\dagger$ denotes the results reproduced by our implementation based on released source code, and $\textbf{DeC}$ denotes our proposed decomposition and composition method.
		}
		\label{tab:sofa}
	\end{table*}
	
	\textbf{Baselines}. 
	{We apply our decomposition and composition operations (denoted \textbf{DeC}) to three open-source representative SGG models}: Motifs \cite{zellers2018neural}, GPSN~\cite{Lin_2020_CVPR}, and TransE~\cite{zhang2017visual}, and compare performance with these models without DeC. 
	Besides, we compare with a number of other recent SGG models: KERN~\cite{chen2019knowledge}, GLAT~\cite{zareian2020learning}, GB-Net~\cite{zareian2020bridging}, and TDE~\cite{tang2020unbiased}. For TDE, we choose TransE~\cite{zhang2017visual} as the base model and SUM~\cite{tang2020unbiased} as the fusion function since it was shown that this combination obtains the best performance~\cite{tang2020unbiased}. 
	
	
	\subsection{Implementation Details} \label{sec:imple}
	Following recent works~\cite{zellers2018neural,tang2020unbiased}, we use the pre-trained Faster-RCNN~\cite{ren2015faster} as the object detection network and freeze its parameters for training our SGG models. For each batch, we set the number of predicates $N=5$ and number of images per predicate $K=1$. During training, we do not generate all visual components at once, given the large total number. Instead, we  maintain a fixed-size visual component dictionary, whose size is set to $3,000$. When the dictionary is full, we randomly choose components to be removed.
	For the relation classification, we set task-specific maximum number of relation triples for each image: $256$ for \textbf{PredCls} and \textbf{SGCls}. For the task of \textbf{SGDet}, the maximum number of object proposals is set to $80$ and the number of classification triples to $1,024$. Additionally, the total number of training iterations is about $130,000$ and the  number of composed relation triples is bout $600,000$, including inter-  and intra-class composition. 
	

	
	\subsection{Evaluation Tasks and Metrics} 
	
	We evaluate our model on the common task of relation retrieval (\textbf{RR}) and its few-shot (\textbf{FR}) and zero-shot (\textbf{ZR}) variants, which are increasingly difficult due to the reduction in training samples. 
	Specifically, \textbf{FR} tests a model's ability to learn from only a few training examples.  We only sample a few samples (5 and 10) for each predicate to train the model. 
	\textbf{ZR}~\cite{lu2016visual} aims to evaluate a model's generalisability on unseen relations. During the testing stage, we select new relation triples as evaluation triples, following TDE~\cite{tang2020unbiased}. 
	Each task includes three sub-tasks: (a) Predicate Classification (\textbf{PredCls}) with ground-truth object labels and boxes; (b) Scene Graph Classification (\textbf{SGCls}) with only the labels; and (c) Scene Graph Detection (\textbf{SGDet}) with neither of them. 
	
	In this paper, we report the unbiased metric mean Recall@K ($\mathbf{{mR@K}}$) instead of the conventional one $\mathbf{{R@K}}$, as it has been shown in the literature that $\mathbf{{R@K}}$ is biased and does not reflect a model's true performance on tail relations~\cite{chen2019knowledge,tang2020unbiased,wang2020tackling}. {Note that all experiments are under a  constrain scheme~\cite{chen2019knowledge}.}
	

	\begin{figure*}
		\centering
		\includegraphics[width=0.85\linewidth]{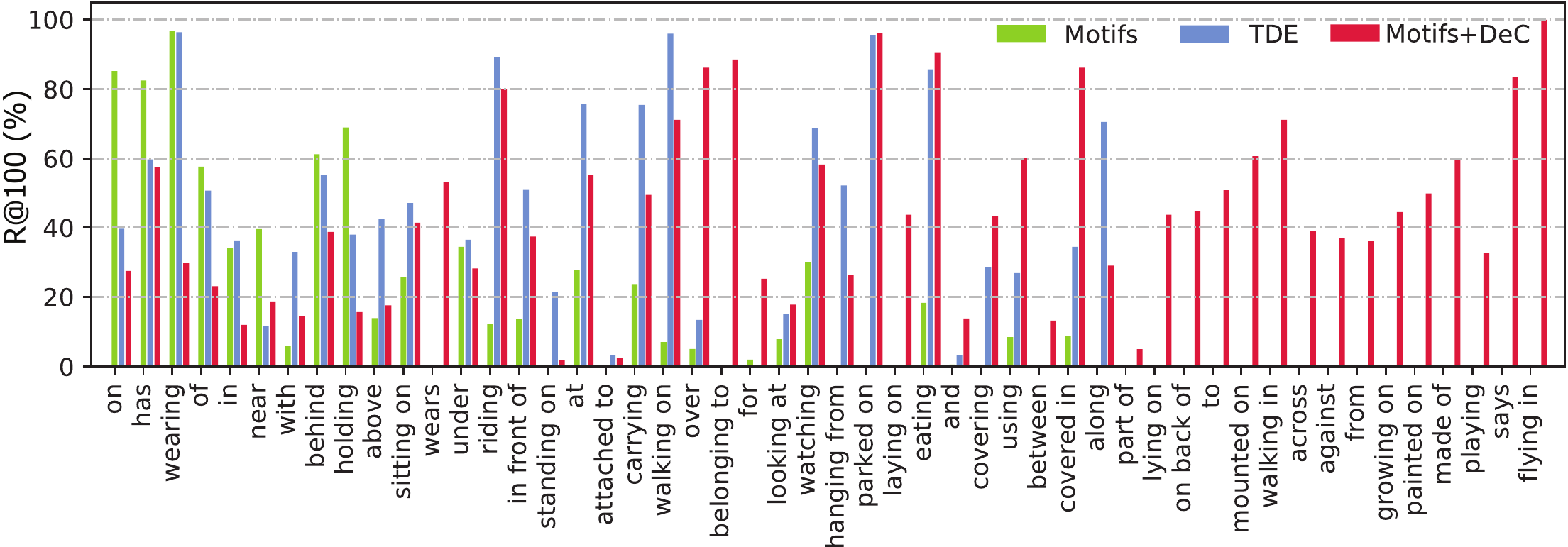}
		\caption{$R@100$ results of PredCls (Constraint) on VG. Note that the frequencies of the relations on the $x$-axis descend from left to right.}
		\label{fig:sp}
	\end{figure*}
	
	\subsection{Relation Retrieval (RR)} 
	Table~\ref{tab:sofa} shows the comparison results with the state-of-the-art methods. It can be clearly observed that when the three baseline models, TransE, GPSN and Motifs, are equipped with our proposed \textbf{DeC} strategy, their performance gains a significant improvement. For instance, for Motifs~\cite{zellers2018neural}, with our proposed $\textbf{DeC}$ added, it obtains approx.\ 2x improvement compared with the original Motifs across all three subtasks, and surpasses the best method TDE by nearly $50$\%. These results clearly demonstrate the effectiveness and generalisability of our \textbf{DeC} strategy.
	
	For a more comprehensive comparison, we further test specific improvement on each predicate, as shown in Figure~\ref{fig:sp}, where we report the results of R@100 of {PredCls} on two representative methods, Motifs and TDE, as the compared methods. The relationships on the $x$-axis are sorted by the numbers of instances in the training set, with the most frequent on the left. From this analysis, we can observe that while our model achieves competitive, albeit slightly worse performance on the head relations, it achieves significantly better performance on the tails relations, without 0 R@100 values on any relation. It can be seen that both Motifs and TDE achieve a near $0$ R@100 value for all of the $15$ least frequent relations (\texttt{part of} onwards). In contrast, our strategy achieves substantially better results, with R@100 scores of at least $40\%$ for $10$ of the $15$ relations, and only one relation (\texttt{part of}) with a R@100 value below $20$. 
	Moreover, for the most challenging relation \texttt{flying in} with only $5$ samples in the training set, our model achieves a 100\% R@100 score, whereas neither Motifs and TDE recognises it (with a R@100 score of $0$). 
	
	Interestingly, we can also find that the performance of our method moderately degrades in some head relations, e.g., \texttt{has} and \texttt{on}, compared with Motifs. We deem that the possible reason is that Motifs adds the frequency bias information in its prediction, and due to the extreme imbalanced distribution of the relations, in which the top-$10$ highly frequent relations take up almost 90\%~\cite{chen2019knowledge} of samples, the added bias knowledge plays a positive role for these head relations. At the same time, this frequency bias is harmful to the tail relations for the same reason. To some extend, we can view it as overfitting on the head relations and underfitting on the tail relations. In contrast, our \textbf{DeC} strategy will supplement those rare relations and relationships and consequently make the model acquire much more balanced performance on all the relationships, thus effectively tackles the long-tail relation classification problem. 
	
	\subsection{Few-shot Relation Retrieval (FR)}
	Few-shot learning for SGG is an important and realistic metric, as the tail relationships have very few samples. To evaluate the performance of \textbf{FR}, we randomly sample $S$ images for each predicate to train the models. More concretely, we select {$S=[5,10]$} as the number of images per predicate for training. We choose  five representative baselines: Motifs, KERN and, GPSN, TransE, and TDE, and Table~\ref{tab:fs} shows the results of FR on the three subtasks, where we also apply our \textbf{DeC} strategy to Motifs, GPSN and TransE. It is worth noting that due to the balanced training set, the statistical bias knowledge proposed in Motifs~\cite{zellers2018neural}  is discarded for all methods. 
	
	From the results it is easy to see the challenging nature of the FR task, where TDE achieves competitive results on the three tasks. When we add \textbf{DeC} to the three models, it can be clearly observed that the three methods gain results comparable to and better than TDE, even though their original versions are much worse than TDE. For instance, after plugging \textbf{DeC} to Motifs,  we can find its performance is on average about $2$ points higher than TDE in PredCls. Additionally, we can observe that both of the original Motifs and KERN struggle in FR, possibly because both of them highly depend on the biased prior knowledge, which becomes ineffective in the few-shot setup. 
	
	\begin{table}[ht]
		\centering
		\resizebox{0.99\columnwidth}{!}{
			\begin{tabular}{llccc}
				\toprule
				\multirow{15}{*}{\small S=5} &   \multirow{2.5}{*}{Method} & SGDet & SGCls & PredCls \\ \cmidrule{3-5} 
				& & mR@50/100 & mR@50/100 & mR@50/100 \\ \cmidrule{1-5}

				& KERN~\cite{chen2019knowledge} & 1.8 / 3.2  & 3.0 / 4.5  &  7.1 / 8.4 \\ 
				
				& TDE~\cite{tang2020unbiased} & 2.1 / 3.3  &  4.2 / 5.2 & 9.1 / 11.7 \\  \cmidrule{2-5}
				& TransE~\cite{zhang2017visual} & 0.7 / 1.2 & 2.2 / 3.4  &  4.6 / 6.5 \\ 
				
				& TranE+\textbf{DeC} & {2.2 / 3.5  }& {3.2 / 4.7} & { 8.3 / 11.2} \\ \cmidrule{2-5}
				& GPSN~\cite{Lin_2020_CVPR} & 1.0 / 2.1 &  2.8 / 4.4  &  8.3 / 10.3 \\
				& GPSN+\textbf{DeC} & {2.7 / \textbf{4.5}  }& \textbf{{6.3 / 7.5}} & { 10.8 / 13.0} \\ \cmidrule{2-5}
				& Motifs~\cite{zellers2018neural} & 1.4 / 2.7 & 2.7 / 4.0  &  7.6 / 9.5 \\ 
				& Motifs+\textbf{Dec} & {\textbf{3.5} / 4.2  }& {6.0 / 7.1} & { \textbf{12.6 / 13.7}} \\ 
				\midrule \midrule
				
				\multirow{10}{*}{\small S=10}
				
				& KERN~\cite{chen2019knowledge} & 2.2 / 4.1  & 5.0 / 6.3 & 10.1 / 12.5 \\ 
				
				& TDE~\cite{tang2020unbiased} & 2.7 / 3.9  &  5.3 / 6.8 & 11.5 / 13.4 \\ \cmidrule{2-5}

				& TransE~\cite{zhang2017visual} & 1.9 / 3.1 & 4.0 / 5.7  &   8.4 / 11.6 \\ 
				& TransE +\textbf{DeC} & {2.3 / 4.6} &  {5.5 / 7.4}  &  {10.2 / 13.4} \\\cmidrule{2-5}

				& GPSN~\cite{Lin_2020_CVPR} & 2.5 / 4.6 & 4.8 / 6.0  &  10.8 / 12.4 \\
				& GPSN+{DeC} & {3.0 / 5.2} &  {6.1 / \textbf{8.3}}  &  {\textbf{13.7} / 14.8} \\\cmidrule{2-5}

				& Motifs~\cite{zellers2018neural} & 1.8 / 3.3 & 4.5 / 5.8  &   ~~9.6 / 11.9\\ 
				
				& Motifs+\textbf{DeC} & \textbf{3.6 / 5.0} &  \textbf{6.5} / 8.2  &  13.4 / \textbf{15.1} \\ \midrule
				
			\end{tabular}
		}
		\caption{The results of few-shot relation retrieval (FR) on VG.}
		\label{tab:fs}
	\end{table}


	\subsection{Zero-shot Relation Retrieval (ZR)}
	Table \ref{tab:zs} shows the comparison results on the task of \textbf{ZR}. Following the settings in TDE~\cite{tang2020unbiased}, we select unseen relation triples of the test dataset as the evaluated samples.  It is worthing noting that the statistical prior knowledge utilised by Motifs and KERN does not cover unseen relations of the test set, thus is ineffectual for the classification of new triples. In contrast, when we apply our \textbf{DeC} strategy to the three methods, we can clearly observe considerable improvements, especially on GPSN, which enjoys the largest performance lift. Even though on the sub-task of PredCls, TDE achieves the best performance in terms of R@100, TDE shows deteriorated performance on the other two sub-tasks.

	\subsection{Ablation Study}
	We further study the effectiveness of each module in our framework for the three tasks: \textbf{RR}, \textbf{FR} and \textbf{ZR}. Specifically, we divide our framework into five variant: (1) baseline without any our proposed strategy; (2) +inter, denoting the baseline model with the inter-class composition; (3) +intra, denoting that the baseline model with the intra-class composition; (4) +\textbf{DeC}$^{'}$, denoting the baseline with a random strategy to choose $v$ for $u$ in Eq.~\ref{eq.rc} instead of our proposed similarity measurement; and (5) +\textbf{DeC}, which is our full model. We choose Motifs as the baseline model, and the results are shown in Table~\ref{tab:ab}. Note that on the task of FR, we set the $S$=$10$ for each predicate. 
	
	From the results, we could obtain the following observations. 
	On the three tasks, both the inter- and intra-class compositions make positive contributions to model performance, but they have different task-specific effectiveness. More concretely, on the main task of RR. the intra-class composition shows superiority to inter-class, because the majority of relations in the test set consists of common relations, and only a small subset of them are unseen by the training model. Thus, the intra-class composition plays a more essential role on the RR task. As for the task of FR, since all relations have only few samples, the intra-class strategy can compose many extra same relations which leads to a significant improvement. Last, for the ZR task that simply evaluates the unseen relations, the inter-class composition exactly creates novel relation samples, and shows a better effectiveness than the intra-class composition. 
	Addition, when we change the similarity  measurement to a random strategy, we could see that both composition strategies become ineffectual on the three tasks, which confirms that our proposed similarity  calculation method is effective. 

	\begin{table}[]
		\centering
		\resizebox{1\columnwidth}{!}{
			\begin{tabular}{lccc}
				\toprule
				\multirow{2.5}{*}{Method} & SGDet & SGCls & PredCls \\ \cmidrule{2-4} 
				& R@50/100 & R@50/100 & R@50/100 \\ \cmidrule{1-4}

				KERN~\cite{chen2019knowledge}   &  0.7 / 1.3 & 2.6 / 3.4  & 11.4 / 15.1    \\

				TDE~\cite{tang2020unbiased}     & 1.9 / 2.7  & 2.9 / 3.6  & 13.2 / \textbf{17.1}  \\ \midrule
				GPSN~\cite{Lin_2020_CVPR}       & 0.5 / 1.0  &  2.0 / 2.8 &  8.5 / 9.3 \\
				GPSN+\textbf{DeC}       & 1.8 / 3.6  &  2.4 / 3.5 &  10.2 / 13.7 \\\cmidrule{1-4}
				
				TransE~\cite{chen2019knowledge}   &  0.3 / 0.6  & 1.4 / 2.3  & 7.2 /  9.0    \\
				TransE+\textbf{DeC}   &  1.4 / 3.3 & 1.9 / 4.3  & 10.8 / 12.5    \\\cmidrule{1-4}
				
				Motifs~\cite{zellers2018neural} & 0.1 / 0.5  & 2.1 / 3.0  & 10.2 / 14.2 \\ 
				Motifs+\textbf{Dec}              & \textbf{2.9 / 4.0} & \textbf{3.2 / 4.5} & \textbf{ 13.6} / 16.5 \\ \bottomrule
			\end{tabular}
		}
		\caption{The results of zero-shot relation retrieval (\textbf{ZR}) on VG.}
		\label{tab:zs}
	\end{table}  
	\begin{table}[]
		\centering
		\resizebox{1\columnwidth}{!}{
			\begin{tabular}{clccc}
				\toprule
				&  \multirow{2.1}{*}{Method} &SGDet  & SGCls & PredCls \\  
				\multirow{8}{*}{\textbf{RR}}&& mR@50/100 & mR@50/100 & mR@50/100 \\ \cmidrule{1-5}
				&Motifs  & ~~5.7 / 6.8 & 7.2 / 7.8 & 13.5 / 14.8  \\ 
				
				&+inter  & ~~8.4 / 10.1 & 10.3 / 12.5 & 20.4 / 30.2 \\ 
				
				&+intra &  12.2 / 15.0 & 16.5 / 17.9 & 31.4 / 35.9  \\ 
				&+\textbf{DeC}$^{'}$ & 6.2 / 7.4 & 5.7 / 6.2 &  10.1 / 12.6 \\ 
				&+{\textbf{DeC}} & \textbf{13.2} / \textbf{15.6} &  \textbf{18.4} / \textbf{19.1} &  \textbf{35.7} / \textbf{38.6} \\ \midrule

				\multirow{5}{*}{\textbf{FR}} & Motifs &   1.8 / 3.3 & 4.5 / 5.8  &   ~~9.6 / 11.9 \\  
				&+inter  & 1.5 / 2.7 & 4.8 / 5.6 & 10.2 / 12.1 \\ 
				
				&+intra & 3.0 / 4.7 & 6.3 / 8.2 & 12.7 / 13.0 \\ 
				&+\textbf{DeC}$^{'}$ &  0.4 / 1.6 & 1.5 / 2.4 & 4.9 / 5.1 \\ 
				&+{\textbf{DeC}} & \textbf{3.6 / 5.0} &  \textbf{6.5} / \textbf{8.2}  &  13.4 / \textbf{15.1} \\ \midrule
				
				\multirow{5}{*}{\textbf{ZR}} & Motifs  & 0.1 / 0.5  & 2.1 / 3.0  & 10.2 / 14.2 \\  
				
				&+inter  & 2.8 / 3.7 & 3.3 / 4.4 & 13.4 / 16.7 \\ 
				
				&+intra &  0.5 / 1.2  & 2.3 / 3.6  &  ~~9.7 / 14.2 \\ 
				&+\textbf{DeC}$^{'}$ &  0.0 / 0.1 & 1.2 / 2.5 & 5.8 / 6.1\\ 
				
				&+{\textbf{DeC}} & \textbf{2.9 / 4.0} & \textbf{3.2 / 4.8} & \textbf{ 13.6} / 16.5  \\
				\bottomrule

			\end{tabular}
		}
		\caption{The ablation study on the three  tasks: \textbf{RR}, \textbf{FR} and \textbf{ZR}. }
		\label{tab:ab}
	\end{table}

	\section{Conclusion}
	In this paper, we proposed a model-free strategy for scene graph generation, which can strongly tackle the low-shot and zero-shot learning problems in the visual relation detection. 
	Comprehensive experiments strongly demonstrate our model's superiority to current state-of-art methods on relation retrieval, where our model significantly outperforms current best models by near 50\% on average on mR@K. We also achieve the best performance in the more challenging few-shot and zero-shot settings. 
	{\small
		\bibliographystyle{ieee_fullname}
		\bibliography{egpaper_final}
	}
\end{document}